\documentclass[times, review, 10pt]{elsarticle}

\usepackage{amssymb}
\usepackage{framed,multirow}

\usepackage{latexsym}
\usepackage{textcomp,booktabs}
\usepackage{amssymb}
\usepackage{pifont}
\usepackage{url}
\usepackage{cite}
\usepackage{natbib}
\usepackage{xcolor}
\definecolor{newcolor}{rgb}{.8,.349,.1}
\usepackage{marvosym}

\usepackage{amsmath}
\usepackage{hyperref} 
\hypersetup{hidelinks, colorlinks=true, linkcolor=red, citecolor=blue, urlcolor=magenta}

\journal{Pattern Recognition}

\begin{document}

\begin{frontmatter}




\title{PFM-VEPAR: Prompting Foundation Models for RGB-Event Camera based Pedestrian Attribute Recognition }

\author{ Minghe Xu\textsuperscript{1,2}, Rouying Wu\textsuperscript{3}, ChiaWei Chu\textsuperscript{1}, Xiao Wang\textsuperscript{4}, Yu Li\textsuperscript{2}*}  
\address{
1. City University of Macau, Macau SAR, China \\ 
2. Zhuhai College of Science and Technology, Zhuhai, China \\ 
3. Macau University of Science and Technology, Macau SAR, China \\ 
4. School of Computer Science and Technology, Anhui University, Hefei 230601, China \\ 
}


\begin{abstract}
Event-based pedestrian attribute recognition (PAR) leverages motion cues to enhance RGB cameras in low-light and motion-blur scenarios, enabling more accurate inference of attributes like age and emotion. However, existing two-stream multimodal fusion methods introduce significant computational overhead and neglect the valuable guidance from contextual samples. To address these limitations, this paper proposes an Event Prompter. Discarding the computationally expensive auxiliary backbone, this module directly applies extremely lightweight and efficient Discrete Cosine Transform (DCT) and Inverse DCT (IDCT) operations to the event data. This design extracts frequency-domain event features at a minimal computational cost, thereby effectively augmenting the RGB branch. Furthermore, an external memory bank designed to provide rich prior knowledge, combined with modern Hopfield networks, enables associative memory-augmented representation learning. This mechanism effectively mines and leverages global relational knowledge across different samples. Finally, a cross-attention mechanism fuses the RGB and event modalities, followed by feed-forward networks for attribute prediction. Extensive experiments on multiple benchmark datasets fully validate the effectiveness of the proposed RGB-Event PAR framework. The source code of this paper will be released on \url{https://github.com/Event-AHU/OpenPAR}.     
\end{abstract}


\end{frontmatter}


\section{Introduction} 

Pedestrian Attribute Recognition (PAR)~\citep{wangsurvey} aims to describe the key characteristics of pedestrians through a set of attributes, such as long hair, wearing glasses, age, gender, clothing, and shoes. This task plays a crucial role in various practical applications, including intelligent video surveillance~\citep{PROMPTPAR}, person re-identification and retrieval~\citep{huang2024attributeRetrieval}, as well as object detection and tracking~\citep{li2024attmot}.
With the rapid development of deep neural networks, particularly large foundation models~\citep{wang2023PTMSurvey}, the performance of PAR models has been significantly improved in recent years. Nevertheless, PAR remains a challenging task under extreme conditions, such as fast motion, low illumination, and over-exposure.

{By analyzing the technical approaches of existing studies, it can be observed that the majority of algorithms are developed based on RGB cameras. Specifically, Chen et al. proposed VTB~\citep{VTB}, which fuses pedestrian images and attribute phases to achieve high-performance attribute recognition. Wang et al. further introduced PromptPAR~\citep{PROMPTPAR}, which leverages the pre-trained CLIP model through prompting to enable more accurate recognition. Jin et al.~\citep{jinsequencePAR} formulated PAR as a sequence generation problem and developed SeqPAR based on a Transformer decoder network.
Recently, event cameras have attracted increasing attention due to their advantages of low power consumption, high dynamic range, high temporal resolution, and low latency~\citep{gallego2020eventSurvey}. Several studies have incorporated event cameras to assist RGB cameras in enhancing perception under challenging conditions. In particular, RGB-Event fusion has been applied to various tasks, including visual tracking~\citep{zhu2025crsot}, sign language translation~\citep{wang2025EvSLT}, and human activity recognition~\citep{li2025SAFE, wang2025HARDVS}. Moreover, Wang et al. proposed the first benchmark dataset for RGB-Event-based PAR~\citep{EventPAR}. This dataset establishes a solid foundation for future research in this field.}

\begin{figure*}
\centering
\includegraphics[width=1\linewidth]{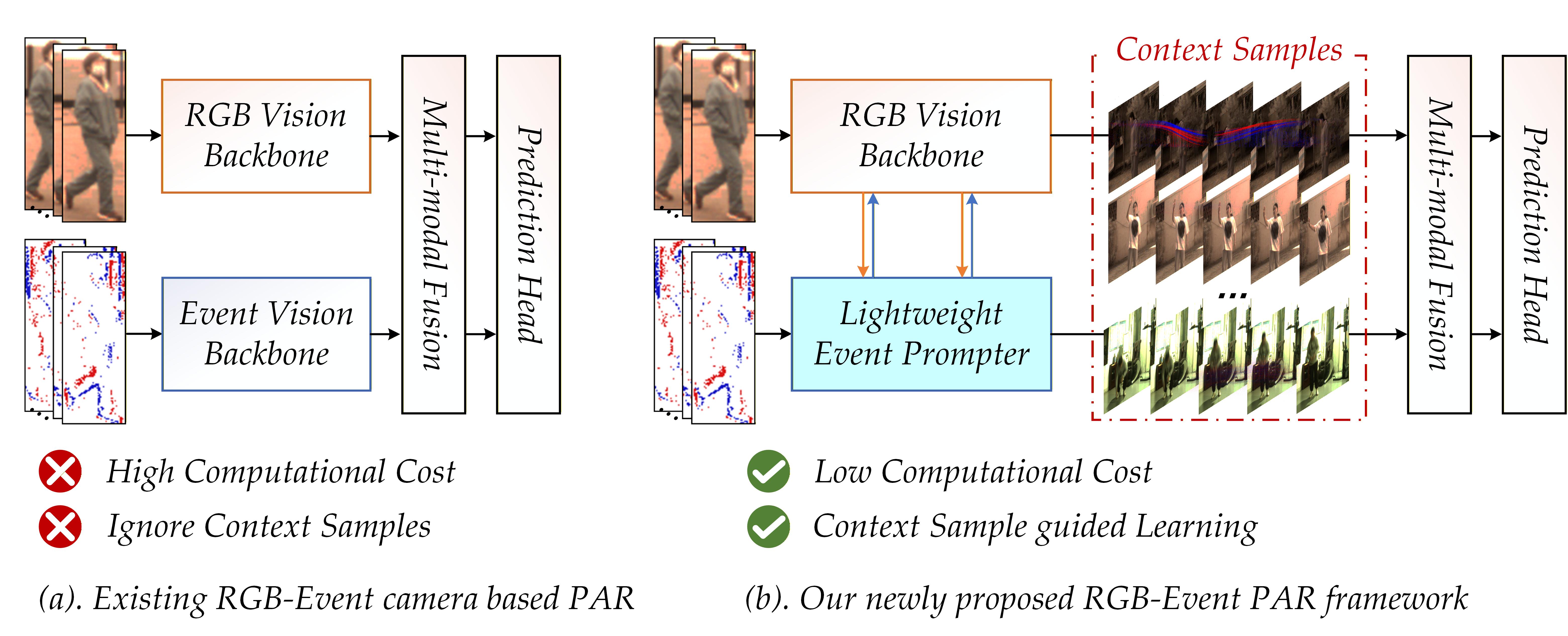}
\caption{Comparison between existing RGB-Event PAR models and the newly proposed one.} 
\label{fig:firstIMG}
\end{figure*}

{Despite these advances, current PAR models still face several limitations:
1). Most existing PAR methods are based solely on RGB cameras. As a result, they often fail under challenging conditions like low illumination, over-exposure, and motion blur.
2). Many approaches treat PAR as a simple image-to-label mapping. They focus heavily on attribute semantics but ignore visual context. This lack of contextual cues often limits overall performance.
3). Current RGB-Event PAR frameworks typically use heavy dual-backbone architectures. They achieve strong results but incur high computational costs. Furthermore, few studies exploit contextual information to further improve performance, as shown in Fig.~\ref{fig:firstIMG}(a). These observations raise a critical question: How can we design a lightweight, context-aware RGB-Event framework? Such a framework must leverage event-based motion cues and rich context to handle challenging visual conditions, all without excessive computational overhead.}

This paper moves beyond the conventional dual-backbone fusion paradigm. Instead, it introduces a novel, lightweight Frequency-aware Event Prompter. This module helps the RGB branch learn more informative features while keeping computational overhead low. Specifically, an input RGB-Event sample is first divided into non-overlapping patches. These patches are then projected and fed into both a ViT-based vision backbone and the  Event Prompter. In addition, a contextual sample mining strategy enhanced by associative memory is proposed. {This approach leverages contextual correlations across different samples, significantly strengthening the model's multi-modal representation capability. Next, a cross-attention mechanism fuses the features from both modalities. Finally, a feed-forward network predicts the pedestrian attribute labels. Fig.~\ref{fig:firstIMG} compares existing RGB-Event PAR frameworks with the proposed approach.}

To sum up, the key contributions of this paper are summarized as follows:

1). This paper moves beyond the conventional dual-backbone paradigm for multi-modal fusion by proposing a lightweight event-stream prompting module. This module effectively assists the RGB visual backbone in extracting and integrating multi-modal features.

2). A context-aware associative memory module, based on the modern Hopfield network, is designed to substantially enhance the visual representation of each sample. It achieves this by establishing associations with diverse contextual instances.

3). Extensive experiments on the EventPAR~\citep{EventPAR} and DukeMTMC-VID-Attribute~\citep{duke} datasets comprehensively validate the effectiveness and generalization capability of the proposed framework for the PAR task.

\section{Related Works} 

In this section, we review related works on pedestrian attribute recognition~\footnote{\url{https://github.com/wangxiao5791509/Pedestrian-Attribute-Recognition-Paper-List}}
, pre-trained foundation models, and modern Hopfield networks.

\subsection{Pedestrian Attribute Recognition}  

{Early research in pedestrian attribute recognition (PAR) evolved from traditional handcrafted features~\citep{HOG} to deep neural networks (DNNs), leading to significant performance improvements.
These deep learning methods can be broadly categorized into two paradigms.
Global-based models process the entire image as input and typically formulate the task as a multi-task learning problem, predicting all attributes simultaneously~\citep{Abdulnabi}.} Although this approach is computationally efficient, it often struggles to capture fine-grained local details.
To address this limitation, researchers have developed local-based models~\citep{8486604} that mimic human perception by combining global features with those extracted from localized regions. Such models often employ body-part detectors or pose estimation techniques to focus on specific regions of interest; however, their performance remains constrained by the accuracy of the underlying localization modules.

To move beyond simple feature extraction, subsequent research has focused on modeling the inherent correlations among attributes and enhancing feature representations.
Sequential models are proposed to capture label dependencies by treating PAR as a sequence generation task~\citep{zhao2019recurrent}.
Other approaches used graph convolutional networks (GCNs) to explicitly model complex relationships among attributes in the form of graph structures~\citep{wang2016cnnrnnunifiedframeworkmultilabel}. This structural correlation is also explored in cross-modal contexts; LS-CAN~\citep{ye2025improving} utilizes a multi-graph correlated neural network to capture the coherence between subjects, attributes, and actions, which provides a new perspective for modeling attribute dependencies.
{Meanwhile, attention mechanisms have been widely adopted to enable models to adaptively focus on the most relevant image regions for each attribute, thereby reducing reliance on rigid, pre-defined localization techniques~\citep{sarafianos2018deep}.}

{Despite increasing model complexity, performance remains heavily dependent on high-quality visual data. To address challenges like low resolution and occlusion, early studies often relied on remedial strategies such as generative models~\citep{croitoru2023diffusion}. In contrast, multi-modal data fusion and prompting strategies offer a more proactive and robust solution for all-weather perception.} 

{For instance, in RGB-T fusion, Lai et al.~\citep{MambaVT} introduced a pure Mamba-based framework to enhance spatio-temporal contextual modeling. Similarly, Xue et al.~\citep{FMTrack} proposed a frequency-aware interaction network combined with multi-expert fusion. To address modality misalignment, Hu et al.~\citep{Hu} designed an efficient dual-branch cross-attention mechanism. Furthermore, for RGB-D data, Tu et al.~\citep{MAPNet} developed a modality-aware prompting network for co-salient object detection. To tackle the inherent noise in depth sensors, Song et al.~\citep{SONG2026113278} proposed a depth correction and edge guidance network.}

{These advanced paradigms have achieved remarkable success in multi-modal fusion. Among various modalities, event cameras have attracted growing interest due to their unique advantages, such as high dynamic range and high temporal resolution. This is exemplified by the explorations of Zhu et al.~\citep{zhu2023visual} in video tracking and Wang et al.~\citep{MvHeat-DET} in object recognition. Different from these works, our proposed RGB-Event PAR framework achieves a better trade-off between computational cost and accuracy.}

\subsection{Pre-trained Foundation Model} 
In recent years, the rise of Foundation Models has sparked a profound paradigm revolution in the field of computer vision~\citep{bommasani2022opportunitiesrisksfoundationmodels}. {Pre-trained on massive datasets~\citep{schuhmann2022laion5bopenlargescaledataset}, these models acquire powerful and generalizable representations. Today, the ``pre-train and fine-tune" paradigm has become mainstream, allowing researchers to rapidly adapt models to downstream tasks with minimal computational cost~\citep{HAP}. }

In Pedestrian Attribute Recognition (PAR), adopting Visual Foundation Models (VFMs) as backbones is now key to boosting performance. For instance, PARFormer~\citep{fan2024parformer} leverages the strong global modeling capability of Vision Transformers (ViT) to capture long-range dependencies. Similarly, Zhou et al.~\citep{zhou2024pedestrian} achieved new SOTA results using ConvNeXt. Meanwhile, Cao et al.~\citep{cao2023novel} demonstrated that the classic ResNet-50 remains highly competitive when combined with feature enhancement.

{Furthermore, foundation models like CLIP, SAM, and MLLMs are being widely extended to more complex visual and cross-modal tasks. In image generation and analysis, they have been successfully applied to real-time image customization~\citep{mao2025realcustom++}, high-fidelity synthesis~\citep{11358752}, and fine-grained image quality assessment using SAM and MLLMs~\citep{SONG2026113420}. Regarding cross-modal and dynamic learning, CLIP has proven effective for cross-modal person re-identification~\citep{XIONG2026113333}. Related techniques have also been extended to emotion-correlated video captioning~\citep{chen2026subjective}. These works collectively demonstrate the immense potential of foundation models in capturing fine-grained features and achieving cross-modal alignment.}

Inspired by these works, we propose in this paper a prompting strategy for pre-trained foundation models to achieve accurate and efficient pedestrian attribute recognition.

\subsection{Modern Hopfield Network}  
{While the classic Hopfield network pioneered auto-associative memory, its low storage capacity and binary nature strictly limit its application. To bridge the gap with modern deep learning, the Modern Hopfield Network (MHN)~\citep{krotov2016dense} was introduced.} MHN exponentially expands storage capacity and supports continuous states, enabling robust content-addressable retrieval for high-dimensional features.The core innovation lies in replacing the quadratic energy function with a new one (e.g., an exponential function), which fundamentally alters the network's dynamics. This modification yields two critical advantages. First, the network's storage capacity is increased from linear to exponential, allowing it to store a vast number of complex patterns without significant interference. Second, the model naturally handles continuous-valued states and patterns, making it perfectly compatible with the feature embeddings widely used in current deep learning models.

Most notably, work by Ramsauer et al. revealed that the update rule of a Modern Hopfield Network is mathematically equivalent to the self-attention mechanism found in the Transformer architecture ~\citep{ramsauer2021hopfieldnetworksneed}. In this interpretation, the ``Query" in attention corresponds to the state pattern of the Hopfield network, the ``Keys" correspond to the stored patterns, and the ``Values" constitute the final output. This equivalence not only provides an energy-based theoretical explanation for the attention mechanism but also highlights the power of Modern Hopfield Networks as highly effective tools for associative memory and pattern retrieval. 

In our work, we leverage this powerful pattern-matching and one-step retrieval capability of the Modern Hopfield Network. We employ it as an external memory module, where the memory bank stores canonical feature prototypes for various pedestrian attributes. {This allows PFM-VEPAR model to query this large-scale knowledge base, enhancing its own feature representations with the retrieved prototypical information, thereby effectively improving recognition accuracy and robustness. }

\section{Methodology} 

\begin{figure*}
\centering
\includegraphics[width=1\linewidth]{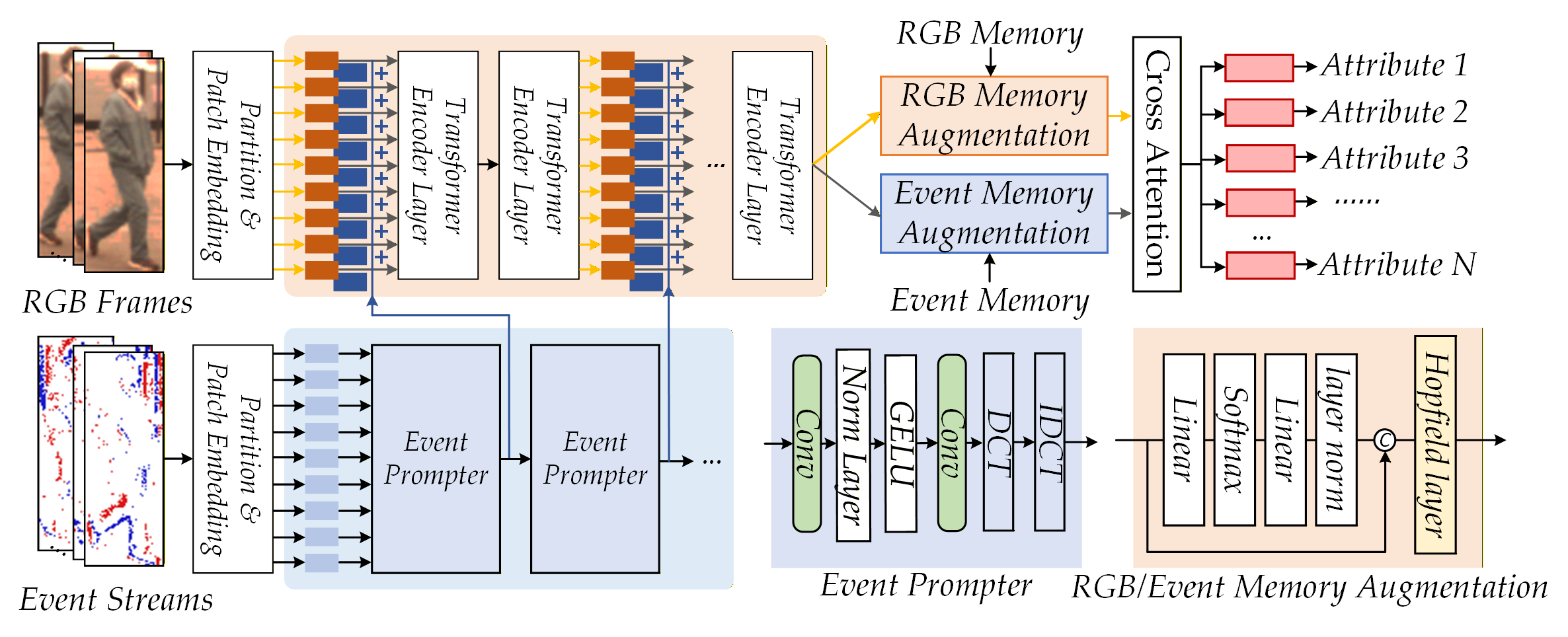}
\caption{An overview of our proposed RGB-Event pedestrian attribute recognition framework, i.e., PFM-VEPAR. In this paper, we propose a lightweight, frequency-aware Event Prompter that enhances RGB feature learning with minimal computational cost. Given an RGB-Event pair, non-overlapping patches are fed into a ViT backbone and the Event Prompter. An associative memory–enhanced context mining strategy strengthens multi-modal representation, followed by cross-attention fusion and a feed-forward network for attribute prediction. }   
\label{fig:framework}
\end{figure*}

\subsection{Overview} 
{This work departs from the conventional dual-backbone architecture commonly used in RGB-Event pedestrian attribute recognition by presenting a lightweight, frequency-aware Event Prompter module.} This module injects motion-rich event cues into the RGB branch in a computationally efficient manner, enabling effective multi-modal interaction without duplicating heavy backbones. {Specifically, given an RGB-Event input pair, the RGB image is first divided into non-overlapping patches. These patches are then linearly embedded and processed by a standard Vision Transformer (ViT) backbone. Concurrently, the corresponding event stream is transformed into frequency-aware prompts via Discrete Cosine Transform (DCT)-based encoding. These prompts are subsequently injected into the intermediate layers of the ViT through the proposed Event Prompter. To further enrich representation learning, an associative memory-enhanced contextual sample mining mechanism, grounded in the Modern Hopfield Network, is introduced. This strategy retrieves and integrates relevant contextual information from neighboring samples, thereby strengthening the semantic and motion-aware features of the current instance.}

Finally, multi-modal features are fused via cross-attention, and attribute predictions are generated using a lightweight feed-forward network. As shown in Fig.~\ref{fig:framework}, the proposed framework offers a more efficient and context-aware alternative to existing RGB-Event PAR pipelines.

\subsection{PFM-VEPAR Framework}  

\noindent $\bullet$ \textbf{Input Representation.~}  
{The input of the proposed model consists of two complementary data modalities.} The first modality is an RGB image, denoted as $I_{rgb} \in \mathbb{R}^{3 \times H \times W}$, where $H$ and $W$ represent the spatial dimensions. This image provides rich color and texture information. The second modality is a spatially and temporally aligned event data stream, represented as $\mathcal{E} = \{e_1, e_2, \dots, e_N\}$. Each event is defined as $e_i = (x, y, t, p)$, where $(x, y)$ denotes the spatial coordinates, $t$ is the timestamp, and $p$ indicates the polarity. This stream records minute brightness changes in the scene, making it highly sensitive to motion edges and dynamic details. To facilitate neural network processing, the asynchronous event stream is first aggregated into a series of temporally contiguous event frames, denoted as $I_{event} \in \mathbb{R}^{F \times 3 \times H \times W}$. Here, $F$ represents the number of event frames in the current sample. This aggregation forms a dense data tensor that serves as the input for subsequent stages.

\noindent $\bullet$ \textbf{RGB Image Encoding.~} 
The RGB image $I_{rgb}$ serves as the primary input pathway for our Vision Transformer backbone. Following the established methodology for processing images in vision transformers, the 2D image is first converted into a 1D sequence of tokens. This is achieved by the PatchEmbed module, which utilizes a convolutional layer to divide $I_{rgb}$ into $N_p$ non-overlapping patches and linearly projects each patch into a $D$-dimensional embedding vector. This operation results in a sequence of patch tokens $T_{rgb} \in \mathbb{R}^{B \times N_p \times D}$. Subsequently, a learnable class token ($T_{cls}$) is prepended to the sequence, and a set of 1D learnable positional embeddings ($E_{pos}$) are added to the combined tokens to provide the model with spatial information. This final sequence, $T_{in} \in \mathbb{R}^{B \times (N_p + 1) \times D}$, is then processed by the stack of standard Transformer encoder blocks, which is where the interaction with the event modality occurs. Many contemporary dual-modal methods employ heavy, dual-stream architectures (e.g., two separate ViT backbones) which incur substantial computational costs. To circumvent this, the Event Prompter was introduced as a lightweight, auxiliary mechanism.

\noindent $\bullet$ \textbf{Event Prompter.~} 
{Dispensing with complex network architectures and discarding the cumbersome full parallel encoder, the Event Prompter serves as an extremely lightweight and efficient alternative. Its core objective is to transform the raw event data stream into a set of high-dimensional semantic prompts at a minimal computational cost. Subsequently, these prompts are utilized to deeply modulate the RGB feature learning process within the single main backbone.} This transformation process begins with a hierarchical convolutional module. This module performs initial feature abstraction on the event frames, generating feature maps with a reduced spatial resolution of $H/4 \times W/4$. {Subsequently, a frequency-domain filtering technique based on the Discrete Cosine Transform (DCT)~\citep{MvHeat-DET} is employed.  This operation effectively suppresses high-frequency noise, thereby enhancing signal purity.}

The purified event feature maps are then tokenized using a ViT patch embedding layer. This operation produces a sequence of raw event tokens, denoted as $T_{event\_raw} \in \mathbb{R}^{(B \cdot F) \times N_p \times D}$. Here, $N_p$ represents the number of patches, and $D$ indicates the embedding dimension. Subsequently, these tokens are temporally aggregated via mean pooling across the $F$ frames. This aggregation yields an average representation, formulated as $\bar{T}_{event} \in \mathbb{R}^{B \times N_p \times D}$. Finally, a simple linear projection layer condenses this sequence into a fixed number of $P$ prompt vectors:

\begin{equation}
T_{prompt} = \text{Proj}(\bar{T}_{event}) \in \mathbb{R}^{B \times P \times D}
\end{equation}

These prompts encapsulate the core dynamics of the event stream. They are dynamically injected into specific layers of the ViT backbone to guide the processing of RGB image tokens. This localized modulation mechanism enables event information to influence the formation of RGB features across various semantic depths. {Crucially, this mechanism achieves cross-modal enhancement without introducing massive additional parameters or permanently altering the backbone's computational graph. Consequently, it minimizes the computational burden of multimodal fusion.}

\noindent $\bullet$ \textbf{RGB/Event Memory Augmentation Module.~}   
Once extracted from the backbone, the deeply fused features enter a sophisticated memory augmentation module. This module performs final refinement and generalization. Its core is a dual-memory system. This system operates synergistically to optimize features from both internal and external perspectives.

\begin{figure*}
\centering
\includegraphics[width=1\linewidth]{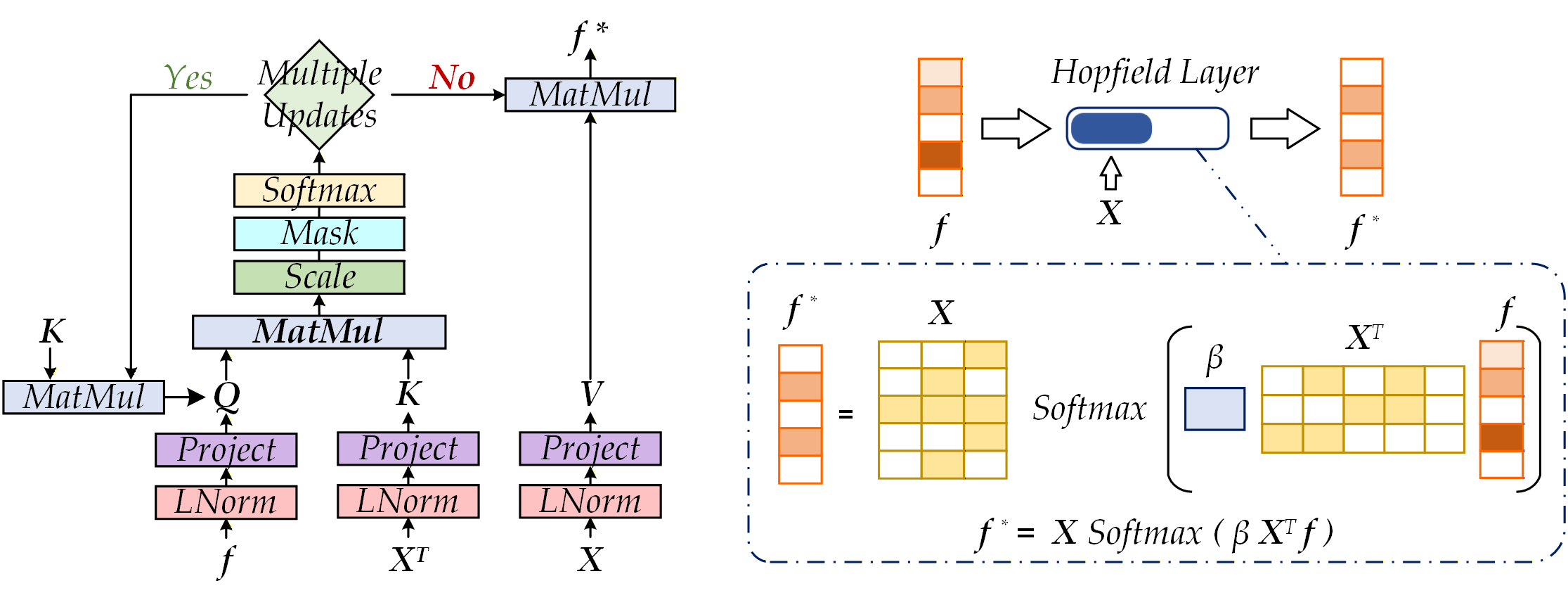}
\caption{An illustration of the (left) Modern Hopfield Network and (right) Hopfield layers.}
\label{fig:MHN}
\end{figure*}

First, the features undergo internal refinement. This is achieved through a self-associative memory mechanism based on the modern Hopfield network~\citep{AMMRG}. This mechanism significantly enhances representation clarity. As illustrated in Fig.~\ref{fig:MHN}, the input feature $X$ is projected into an implicit prototype space to retrieve relevant content. This operation is defined as follows:
\begin{equation}
X' = X + \text{Dropout}(\text{Content}(\text{Softmax}(\beta \cdot X W_{lookup})))
\end{equation}
where $W_{lookup}$ projects $X$ to the prototype space, $W_{content}$ projects it back, and $\beta$ is a scaling temperature, typically $1/\sqrt{dim}$. This reinforces components consistent with core patterns while suppressing irrelevant noise.

Subsequently, the purified features $X'$ serve as query signals ($Q$). The module then interacts with a large-scale external associative memory bank. This memory bank stores a vast collection of ``canonical" feature prototypes, denoted as Keys ($K$) and Values ($V$). {Specifically, to ensure the representativeness of these prototypes and prevent the introduction of inference noise, the external memory bank is statically constructed during the offline initialization phase. First, features are extracted from the training set using a weighted sampling strategy to mitigate the issue of attribute imbalance. Subsequently, K-Means clustering is applied to the collected feature pool for each specific attribute, and 100 cluster centers are extracted as the final representative prototypes. This static and fixed-size memory bank design not only guarantees the diversity and compactness of the feature space but also acts as a natural regularization mechanism, effectively preventing the memory pollution potentially caused by dynamic updates.} Based on the constructed static memory bank, an efficient retrieval, mathematically equivalent to an attention operation, is performed:
\begin{equation}
M_{retrieved} = \text{softmax}(\beta \cdot Q K^T) V
\end{equation}
This allows the model to associate and calibrate its own features with more universal prior knowledge. Consequently, the features are effectively calibrated.

An innovative cross-modal consistency-gating mechanism intelligently governs this retrieval process. It assesses the semantic alignment between the retrieved RGB features ($M_{rgb}$) and Event features ($M_{evt}$). This assessment yields a similarity weight. Subsequently, this weight determines an adaptive factor, $\alpha$. This factor dynamically adjusts an additive fusion strategy. Rather than simply adding memory to its corresponding modality (e.g., $X_{rgb} + M_{rgb}$), the model executes a cross-modal memory infusion:
\begin{equation}
X_{rgb\_out} = X_{rgb} + \alpha M_{rgb} + (1-\alpha) M_{evt}
\end{equation}
\begin{equation}
X_{evt\_out} = X_{evt} + (1-\alpha) M_{rgb} + \alpha M_{evt}
\end{equation}

This strategy ensures both effective and robust information enhancement. Finally, the entire augmentation process concludes with a dual-layer, 8-head bidirectional cross-attention mechanism. This mechanism performs a fine-grained alignment on the memory-augmented features ($X_{rgb\_out}$, $X_{evt\_out}$). Consequently, cross-modal information is fully integrated. This integration ultimately produces a high-quality feature representation for classification.

\noindent $\bullet$ \textbf{Attribute Recognition Head.~}    
The refined and enhanced feature representation $F_{final} \in \mathbb{R}^{B \times N \times D_{fused}}$ is passed to a simple classification head. This module first applies a global average pooling (GAP) operation to consolidate the feature map into a fixed-length global feature vector $\bar{f} \in \mathbb{R}^{B \times D_{fused}}$. Following this, a fully-connected layer maps this vector to the attribute space, outputting an independent prediction score (logit) for each attribute. The logits $L$ are computed as:
\begin{equation}
L = \bar{f}W_{cls} + b_{cls}
\end{equation}
where $W_{cls}$ and $b_{cls}$ are the weight and bias of the final linear classifier.

\subsection{Loss Function}   
PAR is a classic multi-label classification task, where the presence or absence of each attribute can be treated as an independent binary classification problem. Accordingly, we employ the Binary Cross-Entropy (BCE) with Logits loss as the fundamental objective function for the model. To address the class imbalance problem prevalent in pedestrian attribute datasets, where certain attributes appear far more frequently than others, we introduce a sample weighting strategy. This strategy dynamically assigns weights to different samples or attributes based on their distribution, thereby encouraging the model to pay more attention to minority classes during training.

Specifically, for the $i$-th sample and the $j$-th attribute in a batch, the weighted BCE loss $L_{ij}$ is calculated as follows:
\begin{equation}
L_{ij} = -w_{ij} \left[ y_{ij} \log(\sigma(p_{ij})) + (1-y_{ij}) \log(1-\sigma(p_{ij})) \right]
\end{equation}

\begin{equation}
w_{ij} = 
\left\{
\begin{array}{ll}
\exp(1 - r_j), & \mbox{if } y_{ij} = 1 \\ 
\exp(r_j), & \mbox{if } y_{ij} = 0 
\end{array}
\right.
\label{eq:weight}
\end{equation}
where $p_{ij}$ is the raw logit output from the model for that attribute, $\sigma(\cdot)$ is the sigmoid function, and $y_{ij} \in \{0, 1\}$ is the ground-truth label. {The term $w_{ij}$ is the weight designed to balance the importance between positive/negative samples and different attributes. Specifically, as shown in Eq. \ref{eq:weight}, it is computed based on the positive sample ratio $r_j$ of the $j$-th attribute in the training set.}

\section{Experiments} 

\subsection{Datasets and Evaluation Metric}  
{Two distinct PAR datasets are utilized to comprehensively evaluate the performance of the PFM-VEPAR framework.} A brief introduction to the two datasets is given below. 

\noindent $\bullet$ \textbf{EventPAR~\citep{EventPAR}} dataset is specifically used to test the model's performance and robustness on data captured by novel sensors. The EventPAR dataset is collected using event cameras and contains more than 10,000 pedestrian samples. It is annotated with a comprehensive set of 50 distinct attributes. This makes it an essential benchmark for validating algorithmic performance in unconventional and challenging conditions, such as environments with high dynamic range or poor illumination.

\noindent $\bullet$ \textbf{DukeMTMC-VID-Attribute~\citep{duke}} is an extension of the DukeMTMC-VID dataset, adapted for attribute recognition tasks. It is particularly valuable for evaluating models in realistic and challenging environments. The dataset's multi-label attributes are decomposed into 36 binary attributes. The training split contains 16,522 images of 702 unique pedestrians, while the testing split consists of 17,661 images of another 702 pedestrians.

{To evaluate the proposed model, five standard metrics for multi-label classification are adopted. The primary metric is mean Accuracy (mA), which provides a balanced measure across all attributes. To provide a more detailed performance analysis, instance-based Accuracy, Precision, Recall, and F1-score are also reported. These metrics are derived from the number of True Positives (TP), True Negatives (TN), False Positives (FP), and False Negatives (FN), and are formulated as follows:}

\begin{equation}
    \mathrm{mA} = \frac{1}{C} \sum_{i=1}^{C} \frac{TP_i + TN_i}{TP_i + TN_i + FP_i + FN_i}
\end{equation}
\begin{equation}
    \mathrm{Acc} = \frac{\sum_{i=1}^{C} (TP_i + TN_i)}{\sum_{i=1}^{C} (TP_i + TN_i + FP_i + FN_i)}
\end{equation}
\begin{equation}
\mathrm{Precision} = \frac{\sum_{i=1}^{C} TP_i}{\sum_{i=1}^{C} (TP_i + FP_i)}, \quad \mathrm{Recall} = \frac{\sum_{i=1}^{C} TP_i}{\sum_{i=1}^{C} (TP_i + FN_i)}
\end{equation}
\begin{equation}
    \mbox{F1-score} = 2 \times \frac{\mathrm{Precision} \times \mathrm{Recall}}{\mathrm{Precision} + \mathrm{Recall}}
\end{equation}

\subsection{Implementation Details}  
During the training and inference stages, the model's input includes one static RGB image and a corresponding stream of five event frames. All inputs are first uniformly resized to a resolution of $256 \times 192$ pixels. To improve the model's generalization, the standard data augmentation techniques are applied to the training images. For model optimization, the AdamW optimizer is selected, with the weight decay set to $5 \times 10^{-4}$.  A warm-up strategy and a cosine annealing learning rate scheduler are employed. The initial learning rate is set to $1.5 \times 10^{-4}$ for both the pre-trained backbone parameters and the new components. The model is trained for 30 epochs, and the batch size is 16. To ensure training stability, gradient clipping is enabled. Additionally, an Exponential Moving Average (EMA) strategy is applied to the model weights to enhance final performance.

Regarding the specific configuration of our proposed modules, the memory augmentation section consists of an internal Hopfield enhancement layer with 1000 prototypes. It also includes an external memory Hopfield module. This external module retrieves information from an offline-built memory bank and uses an additive fusion with a cross-modal interaction strategy to enhance the final features. The proposed model is implemented using the PyTorch deep learning framework, and all experiments are conducted on a server equipped with an NVIDIA RTX 4090D GPU with 24GB of memory. For the visual backbone, we use a Vision Transformer Base (ViT-B) model. For pre-training, half of the dataset is utilized, selected via the $CFS_list.pkl$ list from the TransReID-SSL project. More details are available in the provided source code.

\subsection{Comparison on Public Benchmarks} 
This section comprehensively validates the effectiveness of the proposed PFM-VEPAR model. A detailed performance comparison is conducted against various existing pedestrian attribute recognition algorithms. The evaluation utilizes two public benchmark datasets: EventPAR and DukeMTMC-VID-Attribute. To ensure a fair and complete evaluation of the proposed dual-modal architecture, corresponding event data was simulated for the DukeMTMC-VID-Attribute dataset. This simulation was necessary because the original dataset provided only the RGB modality.

\noindent $\bullet$ \textbf{Results on EventPAR Dataset.~}  Table~\ref{tab:sota_comparison} presents a detailed performance comparison of the PFM-VEPAR model, against various SOTA methods on the EventPAR dataset. The results indicate that our model achieves SOTA performance on key metrics. Specifically, PFM-VEPAR reaches a mean Accuracy (mA) of 90.05\%, which significantly surpasses previous leading methods such as VTB~\citep{VTB} (88.41\%) and HAP~\citep{HAP} (88.28\%). Furthermore, PFM-VEPAR also ranks first in Recall with a score of 90.18\% and remains highly competitive in Accuracy (Acc) and F1-score.

In addition to its excellent accuracy, PFM-VEPAR demonstrates a significant advantage in terms of inference speed. With a test time of only 92 seconds, PFM-VEPAR is considerably faster than other high-performing models, such as OTN-RWKV~\citep{EventPAR} (361s) and HAP~\citep{HAP} (148s). {Overall, these results demonstrate that PFM-VEPAR establishes a new SOTA in recognition accuracy. Furthermore, the model strikes an excellent balance between performance and computational efficiency.}

\begin{table*}[ht]
  \centering
  \caption{Comparison with public methods on EventPAR datasets. {Test time refers to inference time. Size denotes the storage size of the saved model checkpoint file on disk, and \#MS represents the peak GPU memory footprint required during the training optimization process.}}
  \label{tab:sota_comparison}
  \resizebox{\textwidth}{!}{%
  \begin{tabular}{l | l | c c c c c | c | c | c | c | c}
    \toprule
    \textbf{Methods} & \textbf{Publish} & \textbf{mA} & \textbf{Acc} & \textbf{Prec} & \textbf{Recall} & \textbf{F1} & \textbf{Test Time} & \textbf{Params (MB)} & \textbf{Flops (GB)} & \textbf{Size (MB)} & \textbf{\#MS (MB)} \\
    \midrule
    \#01 DeepMAR~\citep{deepmar} & ACPR15 & 66.57 & 69.53 & 74.90 & 88.54 & 81.57 & 64s & 23 & 25 & 181 & 592 \\
    \#02 ALM~\citep{tang2019improving} & ICCV19 & 57.18 & 64.17 & 75.59 & 73.20 & 74.38 & 813s & 17 & 8 & 66 & 785 \\
    \#03 Strong Baseline~\citep{rethinkingofpar} & arXiv20 & 73.75 & 61.86 & 67.23 & 80.78 & 75.43 & 47s & 24 & 25 & 91 & 576 \\
    \#04 RethinkingPAR~\citep{rethinkingofpar} & arXiv20 & 81.37 & 80.84 & 86.31 & 87.57 & 86.93 & 44s & 24 & 24 & 91 & 580 \\
    \#05 SSCNet~\citep{jia2021spatial} & ICCV21 & 63.10 & 66.07 & 72.72 & 83.22 & 77.62 & 49s & 24 & 9 & 90 & 588 \\
    \#06 VTB~\citep{VTB} & TCSVT22 & 88.41 & 83.83 & 87.89 & 89.31 & 88.53 & 243s & 93 & 66 & 333 & 525 \\
    \#07 Label2Label~\citep{label2label} & ECCV22 & 72.49 & 74.01 & 86.60 & 79.02 & 82.19 & 398s & 66 & 37 & 829 & 650 \\
    \#08 DFDT~\citep{zheng2023diverse} & EAAI22 & 61.71 & 63.14 & 79.17 & 70.63 & 74.66 & 287s & 88 & 37 & 669 & 2584 \\
    \#09 Zhou et al.~\citep{zhou2023} & JICAI23 & 56.46 & 60.89 & 73.37 & 73.62 & 73.50 & 42s & 234 & 24 & 536 & 701 \\
    \#10 PARFormer~\citep{fan2023parformer} & TCSVT23 & 83.12 & 80.48 & 85.14 & 88.41 & 86.53 & 438s & 195 & 205 & 756 & 2481 \\
    \#11 SequencePAR~\citep{jin2023sequencePAR} & arXiv23 & 86.27 & 84.42 & 88.81 & 89.12 & 88.83 & 1497s & 466 & 577 & 1776 & 9901 \\
    \#12 VTB-PLIP~\citep{zuo2024plip} & arXiv23 & 67.25 & 68.37 & 77.75 & 79.72 & 78.37 & 229s & 31 & 591 & 591 & 501 \\
    \#13 Rethink-PLIP~\citep{zuo2024plip} & arXiv23 & 68.75 & 70.03 & 81.82 & 78.04 & 79.89 & 37s & 21 & 23 & 144 & 481 \\
    \#14 PromptPAR~\citep{PROMPTPAR} & TCSVT24 & 86.51 & 82.27 & 86.35 & 89.36 & 87.64 & 1312s & 8 & 5 & 1200 & 2296 \\
    \#15 HAP~\citep{HAP} & NIPS24 & 88.28 & 85.42 & 89.54 & 89.72 & 89.63 & 148s & 86 & 99 & 327 & 894 \\
    \#16 SSPNet~\citep{sspnet} & PR24 & 66.92 & 67.49 & 78.73 & 76.90 & 77.80 & 69s & 40 & 26 & 100 & 1898 \\
    \#17 MambaPAR~\citep{mambapar} & arXiv24 & 50.01 & 42.32 & 54.81 & 57.31 & 55.63 & 105s & 0.64 & 0.41 & 98 & 802 \\
    \#18 MaHDFT~\citep{MaHDFT} & arXiv24 & 50.43 & 44.98 & 59.10 & 59.70 & 58.57 & 292s & 164 & 6033 & 499 & 1313 \\
    \#19 OTN-RWKV~\citep{EventPAR} & arXiv25 & 87.70 & 84.94 & 89.15 & 89.48 & 89.18 & 361s & 201 & 77 & 836 & 2413 \\
    \midrule
    PFM-VEPAR (Ours) & - & \textbf{90.05} & \textbf{84.67} & \textbf{88.26} & \textbf{90.18} & \textbf{89.21} & 92s & 108 & 109 & 327 & 824 \\
    \bottomrule
  \end{tabular}%
  }
\end{table*}


\noindent $\bullet$ \textbf{Results on DUKE Dataset.~} To further validate generalization ability and effectiveness, evaluations were conducted on the DukeMTMC-VID-Attribute dataset. The proposed model is compared with other recent methods, as shown in Table~\ref{tab:sota_comparison_video_duke}. The experimental results indicate that the PFM-VEPAR model demonstrates highly competitive performance on this dataset.

Specifically, the model achieves an Accuracy of 67.58\% and a Precision of 80.21\%. Both metrics rank first among all compared methods. Regarding the F1-score, the proposed approach reaches 79.16\%. This slightly surpasses the strong performer, SSPNet (79.15\%). The Recall (78.13\%) is slightly lower than that of SSPNet (79.78\%). Nevertheless, PFM-VEPAR achieves the best or near-best overall results across three key metrics: Accuracy, Precision, and F1-score.These findings fully demonstrate the effectiveness of the architectural design. Furthermore, they highlight a strong generalization capability across different scenarios.

\begin{table}[ht]
  \centering
  \caption{Results on DukeMTMC-VID-Attribute RGB-Event based PAR dataset.}
  \label{tab:sota_comparison_video_duke}
  \begin{tabular}{l | l | c c c c}
    \toprule
    \textbf{Methods} & \textbf{Backbone} & \textbf{Acc} & \textbf{Prec} & \textbf{Recall} & \textbf{F1} \\
    \midrule
    Zhou et al.~\citep{zhou2023} & ConvNext & 65.51 & 78.42 & 77.40 & 77.91 \\
    VTB-PLIP~\citep{zuo2024plip} & ResNet50 & 43.95 & 60.80 & 58.52 & 59.10 \\
    Rethink-PLIP~\citep{zuo2024plip} & ResNet50 & 40.25 & 55.41 & 56.13 & 55.77 \\
    SSPNet~\citep{sspnet} & ResNet50 & 67.53 & 78.54 & 79.78 & 79.15 \\
    MambaPAR~\citep{mambapar} & Vim & 61.55 & 75.54 & 74.84 & 74.24 \\
    \midrule
    PFM-VEPAR(Ours) & VTB-B/16 & \textbf{67.58} & \textbf{80.21} & 78.13 & \textbf{79.16} \\
    \bottomrule
  \end{tabular}
\end{table}

\subsection{Component Analysis}  
\begin{figure}
\centering
\includegraphics[width=0.5\columnwidth]{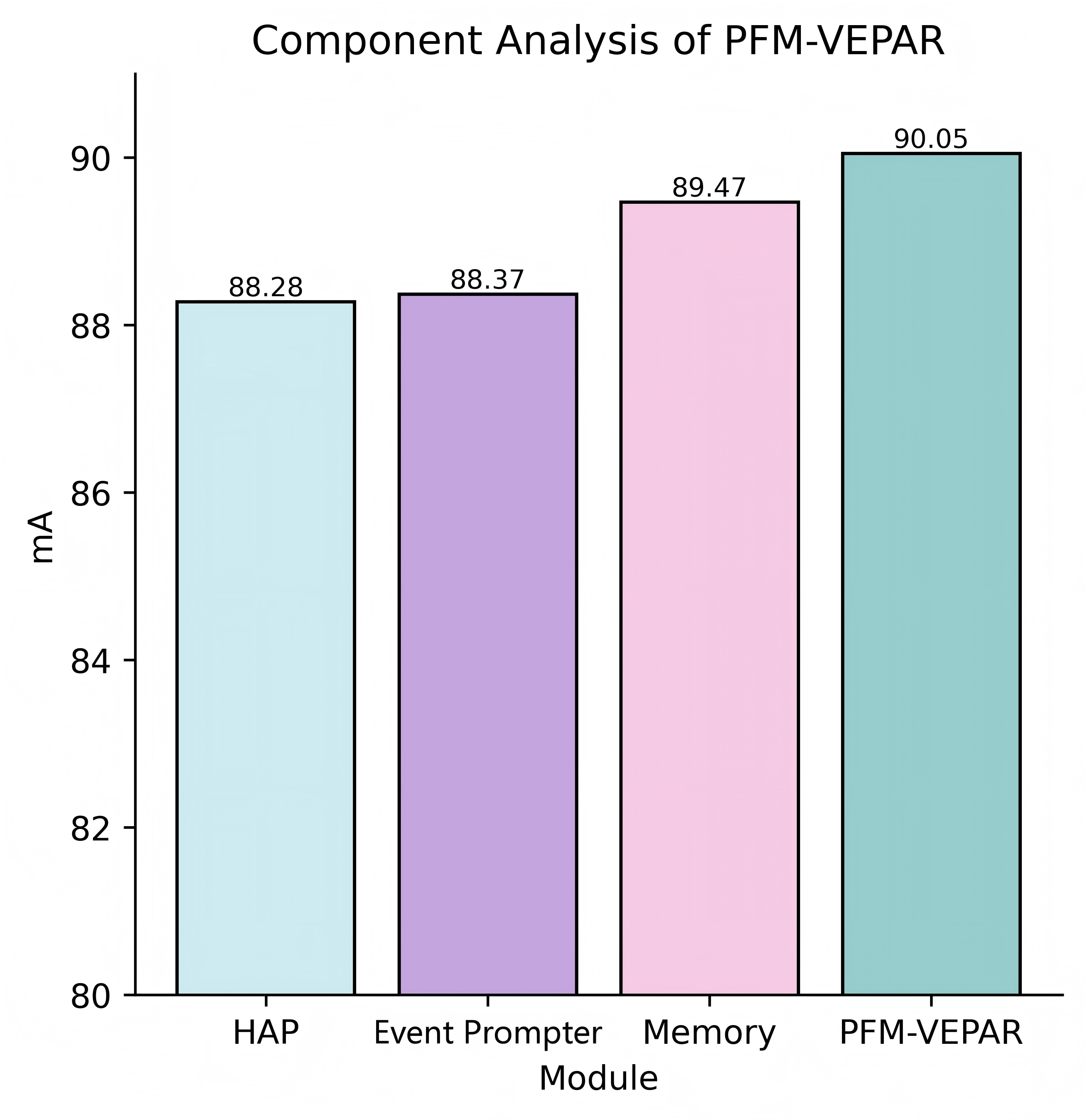}
\caption{Component Analysis of PFM-VEPAR.}
\label{fig:ca}
\end{figure}

To validate the effectiveness of the core components within the PFM-VEPAR model, a detailed component analysis was conducted. The corresponding results are presented in Fig.~\ref{fig:ca}. A strong baseline model, HAP~\citep{HAP}, serves as the starting point for comparison. This baseline achieves an mA of 88.28\%. The experimental data clearly demonstrate that each proposed component uniquely enhances the overall performance.

\begin{enumerate}
\item \textbf{Event Prompter}: When the Event Prompter module is introduced, the model's mA shows a slight increase to 88.37\%. Although there is a dip in Accuracy and Precision, this indicates that the Event Prompter, as a modal fusion mechanism, is beginning to provide a gain in mean performance.
\item \textbf{Memory}: When the memory augmentation module is integrated, the model's mA shows a significant leap to 89.47\%. This provides strong evidence for the substantial benefit of our dual-memory system (internal Hopfield enhancement and external memory retrieval) in refining features and introducing prior knowledge.
\end{enumerate}

Most importantly, when all modules are combined to form the complete PFM-VEPAR model, the performance on the primary metric, mA, reaches the peak of 90.05\%. This result is not only substantially higher than the 88.28\% baseline but also surpasses the performance achievable by either component alone. This clearly demonstrates a powerful synergistic effect between the Event Prompter and the Memory module.

{Furthermore, results indicate that while PFM-VEPAR achieves the best performance in the crucial mA metric and the highest Recall (90.18\%), its Accuracy and Precision are slightly lower compared to the ``Memory-only'' configuration. This is interpreted as a favorable performance trade-off. The significant gain in mA' implies that the proposed model is superior at handling long-tail distributions and recognizing difficult attributes. Concurrently, the boost in Recall' demonstrates that the model is more comprehensive in identifying all positive attributes. Ultimately, this strategy—exchanging a minor drop in `Acc' and `Prec' for significant gains in the critical metrics of `mA' and `Recall'—proves highly valuable for complex real-world applications such as pedestrian attribute recognition.}

\subsection{Ablation Study} 
This section presents a series of detailed ablation studies. {These studies systematically analyze the effectiveness of various components and key parameters within the proposed PFM-VEPAR model. The primary objective is to clearly demonstrate the specific contribution of each module to the final performance. Furthermore, this provides empirical support for the architectural design choices.}

The analysis focuses on three core aspects. First, the impact of different combinations of RGB and event frames is investigated. This step identifies the optimal input data ratio. Second, the configuration of the Event Prompter module is explored. Prompts are injected at various depths of the backbone network. The resulting effects on feature learning are then analyzed to determine the optimal setup. Finally, the memory augmentation mechanism is evaluated. The number of samples used to construct the external memory bank is varied. This examines the influence of memory size on overall performance and robustness.

\noindent $\bullet$ \textbf{Analysis of Input Frames.~}  
{To determine the optimal input combination for the RGB and Event modalities, an ablation study was conducted, fixing the event stream to 5 frames while varying the number of RGB images from 1 to 5.} The results are presented in Table \ref{tab:ablation_frames}. The data clearly shows that the model achieves its best performance with an input of 1 RGB frame and 5 event frames, reaching a mean Accuracy (mA) of 90.05\%. When the RGB input is increased from 2 to 5 frames, the model's performance fluctuated slightly but never surpassed the result of using a single RGB frame. For example, using 4 RGB frames resulted in an mA of 89.95\%, which is very close to but still below the peak performance. This result indicates that for our model, a single, well-chosen RGB image provides sufficient static appearance information to effectively complement the dynamic information captured by the 5-frame event stream. Adding more RGB images may introduce redundant information without significantly enhancing the model's final discriminative capability. {Therefore, based on this experimental analysis, this model selected the configuration of 1 RGB frame and 5 event frames as the final input setting.}

\begin{table}
  \centering
  \caption{Comparison of different input settings.}
  \label{tab:ablation_frames}
  \begin{tabular}{c c | c | c | c | c | c}
    \toprule
    \textbf{RGB} & \textbf{Event} & \textbf{mA} & \textbf{Acc} & \textbf{Prec} & \textbf{Recall} & \textbf{F1} \\
    \midrule
    1 & 5 & 90.05 & 84.67 & 88.26 & 90.18 & 89.21 \\
    2 & 5 & 89.34 & 84.32 & 87.99 & 90.03 & 89.00 \\
    3 & 5 & 89.63 & 84.83 & 88.32 & 90.33 & 89.31 \\
    4 & 5 & 89.95 & 84.58 & 88.23 & 90.09 & 89.15 \\
    5 & 5 & 89.70 & 84.58 & 88.12 & 90.21 & 89.15 \\
    \bottomrule
  \end{tabular} 
\end{table}

\noindent $\bullet$ \textbf{Analysis of Different Localizations and Layers of Event Prompter.~} 
To isolate and evaluate the configuration of the Event Prompter module, a series of ablation studies was conducted excluding the memory augmentation module. Specifically, the frequency transformation method (DCT/IDCT vs. DFT/IDFT) and the layer locations for prompt injection were investigated, with the results presented in Table~\ref{tab:ablation_transform_layers}.

{The experimental results clearly indicate that the model achieves peak performance, with a mean Accuracy (mA) of 88.43\%, when utilizing the DCT/IDCT transform and injecting event prompts at layers 6 and 8. First, the choice of DCT/IDCT over other frequency-domain methods, such as DFT, for processing asynchronous event streams is theoretically and empirically justified. From a theoretical standpoint, event data inherently captures sparse, high-frequency motion and edge information. For such data, DCT offers distinct advantages: (1) it operates strictly in the real domain, thereby avoiding the computational overhead and complex-number representations required by DFT; (2) it exhibits superior energy compaction properties, effectively concentrating critical structural information into fewer coefficients; and (3) unlike DFT, which assumes periodic boundaries and is prone to spectral leakage, DCT assumes even symmetry, rendering it more robust against boundary artifacts in non-periodic event data. Empirically, DCT/IDCT consistently outperforms DFT/IDFT across all corresponding layer configurations. For instance, with injection at layers ``6 8'', the mA for DCT/IDCT is nearly 0.8 percentage points higher than that for DFT/IDFT (88.43\% vs. 87.66\%). This performance gap validates the theoretical superiority of DCT for extracting frequency-domain features from event prompts.}

{Second, regarding the choice of injection layers, a multi-layer injection strategy proves significantly more effective than a single-layer approach. Performance when injecting prompts at two layers (e.g., ``6 8'' or ``8 10'') is generally superior to single-layer injection (e.g., ``8'' or ``10''). However, increasing the number of injection points to three (``6 8 10'') results in a slight performance degradation, with the mA dropping to 88.17\%. This decline suggests that overly dense prompt injection may interfere with the inherent feature learning process of the backbone network. Ultimately, these findings provide critical empirical guidelines for designing effective injection strategies within the Event Prompter.}

\begin{table}
  \centering
  \caption{Comparison of Different Localizations and Layers of Event prompter.}
  \label{tab:ablation_transform_layers}
  \begin{tabular}{l c | c | c | c | c | c}
    \toprule
    \textbf{Transform} & \textbf{Layers} & \textbf{mA} & \textbf{Acc} & \textbf{Prec} & \textbf{Recall} & \textbf{F1} \\
    \midrule
     DCT/IDCT & 8 & 87.24 & 84.32 & 88.42 & 89.53 & 88.97 \\
     DCT/IDCT & 10 & 87.41 & 84.39 & 88.46 & 89.60 & 89.03 \\
     DCT/IDCT & 6 8 & 88.43 & 85.14 & 88.92 & 90.04 & 89.48 \\
     DCT/IDCT & 8 10 & 88.37 & 85.01 & 88.86 & 89.95 & 89.40 \\
     DCT/IDCT & 6 8 10 & 88.17 & 85.10 & 88.88 & 90.04 & 89.46 \\
     DFT/IDFT & 6 8 & 87.66 & 83.10 & 87.22 & 89.32 & 88.25 \\
     DFT/IDFT & 8 10 & 87.43 & 83.09 & 87.27 & 89.26 & 88.26 \\
     DFT/IDFT & 6 8 10 & 87.47 & 83.04 & 87.16 & 89.29 & 88.22 \\
    \bottomrule
  \end{tabular}
\end{table}

\noindent $\bullet$ \textbf{Analysis of Different Samples of RGB/Event Memory Augmentation.~} 
{To evaluate the memory augmentation module and determine its optimal configuration, an ablation study  was conducted on the modalities and the number of memory samples. Instead of random sampling, we applied a clustering-based prototype selection method on the training set to extract representative feature centers for each attribute. Results are shown in Table~\ref{tab:ablation_memory}.}

{The results clearly show that the dual-modal (RGB and Event) memory bank achieves the best mean Accuracy (mA). Tested with 100, 200, and 500 prototypes per attribute, this setup consistently outperforms unimodal ones, validating our dual-modal design. For single-modality configurations, RGB memory consistently beats Event memory. This indicates that RGB features, rich in appearance details, form a stronger foundation. Meanwhile, event features provide a valuable complementary boost when combined.}

{Interestingly, increasing the number of prototypes does not improve performance. The model peaks at 90.05\% mA with a 100-prototype dual-modal memory bank. Performance slightly drops when the number increases to 200 or 500. This decline is directly tied to our clustering strategy. Specifically, 100 cluster centers are sufficient to capture the core intra-class variance. They provide precise and highly representative prototypes for Hopfield retrieval. Forcing the selection of 200 or 500 prototypes inevitably introduces outliers or redundant features from the distribution margins. This adds noise and degrades retrieval quality. Therefore, our final model uses a dual-modal memory bank with 100 prototypes.}

\begin{table}
  \centering
  \caption{Comparison of different samples of RGB/Event Memory Augmentation.}
  \label{tab:ablation_memory}
  \begin{tabular}{c c c | c | c | c | c | c}
    \toprule
    \textbf{Images} & \textbf{RGB} & \textbf{Event} & \textbf{mA} & \textbf{Acc} & \textbf{Prec} & \textbf{Recall} & \textbf{F1} \\
    \midrule
    100 & \checkmark & \checkmark & 90.05 & 84.67 & 88.26 & 90.18 & 89.21 \\
    100 & \checkmark &            & 89.42 & 84.53 & 88.20 & 90.10 & 89.14 \\
    100 &            & \checkmark & 88.23 & 84.47 & 88.43 & 89.75 & 89.08 \\
    \hline 
    200 & \checkmark & \checkmark & 89.42 & 84.51 & 88.09 & 90.18 & 89.12 \\
    200 & \checkmark &            & 88.89 & 84.40 & 88.27 & 89.93 & 89.05 \\
    200 &            & \checkmark & 88.17 & 84.36 & 88.35 & 89.70 & 89.02 \\
    \hline 
    500 & \checkmark & \checkmark & 89.33 & 84.56 & 88.13 & 90.18 & 89.15 \\
    500 & \checkmark &           & 88.59 & 84.56 & 88.10 & 90.25 & 89.16 \\
    500 &           & \checkmark & 88.29 & 84.44 & 88.37 & 89.77 & 89.07 \\
    \bottomrule
  \end{tabular}
\end{table}

\begin{figure*}
\centering
\includegraphics[width=0.9\linewidth]{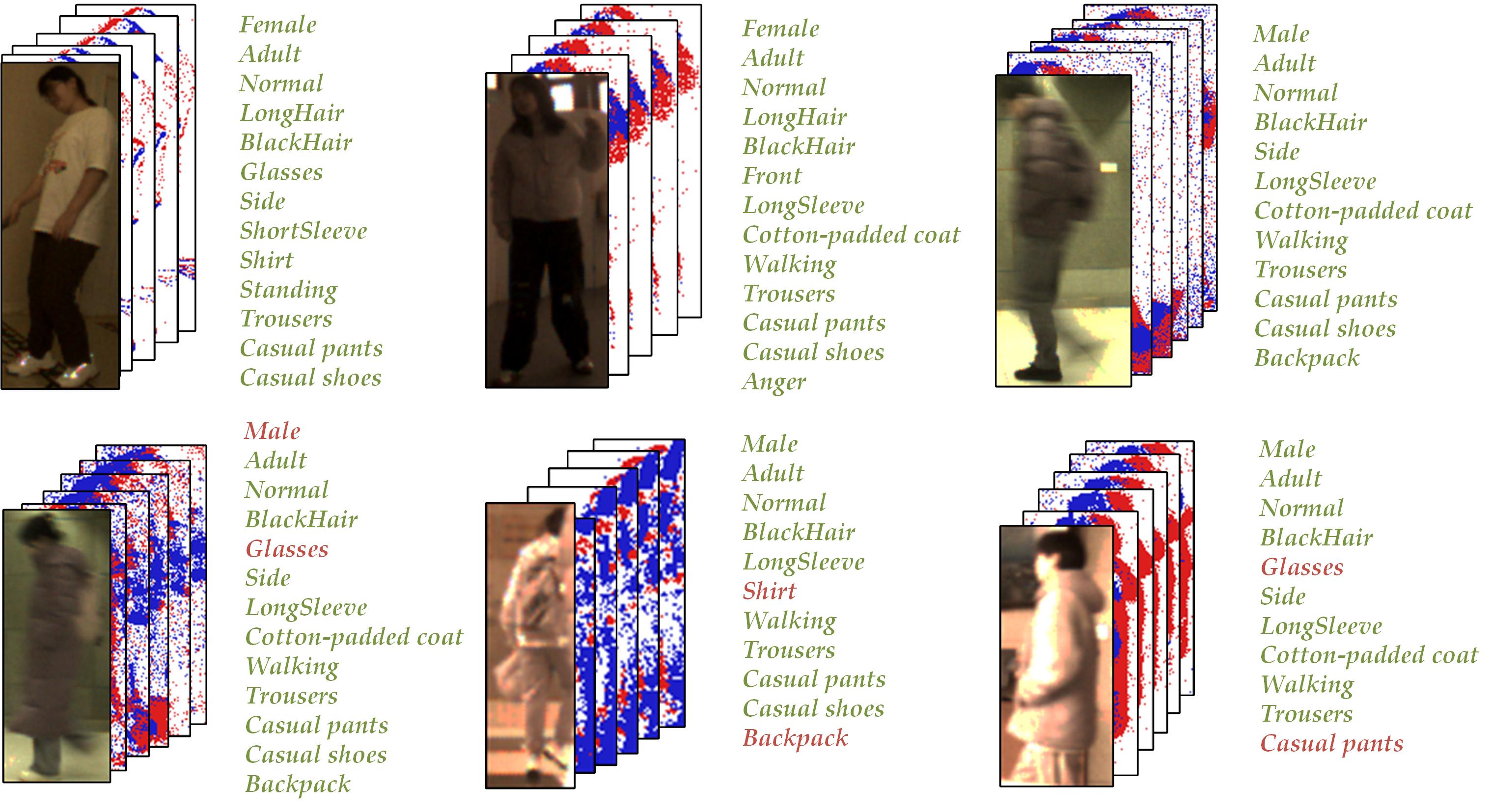}
\caption{Visualization of some predicted pedestrian attributes using our proposed Model PFM-VEPAR.}  
\label{fig:qualitative_results}
\end{figure*}

\subsection{Visualization} 
To intuitively illustrate the performance of the proposed PFM-VEPAR model in real-world scenarios, prediction results for several test samples are visualized. These visualizations are presented in Fig.~\ref{fig:qualitative_results}. A strict confidence threshold is applied to these results. Specifically, a prediction is considered valid only if the model confidence for the corresponding attribute exceeds 90\%. Correctly predicted attributes are marked in green text, while incorrect predictions are highlighted in red. As can be clearly seen in the figure, the PFM-VEPAR model demonstrates strong robustness, even when faced with images of varying clarity, sizes, and significant motion blur. For the vast majority of fundamental attributes, such as `Male', `Adult', `BlackHair', `LongSleeve', and `Walking', the model consistently provides accurate predictions. This indicates that by effectively fusing the static appearance information from RGB with the dynamic edge information from Events, the PFM-VEPAR model has successfully learned discriminative features that are robust to degradations in visual quality. However, the visualizations also reveal that the model still faces challenges when dealing with fine-grained or semantically ambiguous attributes. For example, in some samples, the model incorrectly predicts small-sized object attributes like `Glasses' or `Backpack'. Furthermore, distinguishing between semantically similar categories, such as `Trousers' and `Casual pants', remains difficult. Overall, these qualitative results demonstrate our model's strong capability in recognizing core attributes under complex conditions, while also pointing to future directions for improvement in fine-grained recognition.
    
\subsection{Limitation Analysis}  

Although the PFM-VEPAR model has achieved SOTA performance on multiple benchmarks, there are still several aspects worthy of further exploration in future work:
1). Our qualitative evaluation shows that the model performs robustly on most basic attributes. However, in rare cases, particularly with small objects (e.g., Glasses or Backpack) or semantically similar categories (e.g., different types of Trousers), it may exhibit minor uncertainties. This suggests opportunities for further refinement in capturing extremely subtle visual distinctions. 
2). PFM-VEPAR is a multi-stage framework whose performance, as shown in our ablation studies, is thoughtfully balanced rather than purely maximized by complexity. {A ``less is more" tendency is observed; optimal results are often achieved with fewer input frames, prompter layers, or memory samples, indicating that strategic simplification can enhance generalization.}

\section{Conclusion and Future Works}  

This paper has explored the integration of asynchronous event streams with RGB frames to address critical challenges in pedestrian attribute recognition (PAR), such as pose variations, severe occlusions, and motion blur. To achieve this, a novel dual-modal fusion framework, termed PFM-VEPAR, is proposed. By innovatively integrating a frequency-aware Event Prompter with a Hopfield-based dual-memory enhancement mechanism, the framework effectively extracts dynamic motion cues and leverages dataset-level contextual knowledge. {A major strength of this approach is its ability to perform modality augmentation without the high computational overhead typically associated with traditional two-stream architectures. Consequently, the proposed memory-augmented prompting strategy offers a new, lightweight paradigm for multimodal fusion. Researchers and practitioners in intelligent video surveillance can readily adapt this framework to other event-assisted visual tasks, such as person re-identification or action recognition.}

Extensive empirical evaluations on the EventPAR and DukeMTMC-VID-Attribute datasets demonstrate the superiority of the proposed approach. Specifically, PFM-VEPAR achieves a SOTA mean accuracy (mA) of 90.05\% on the EventPAR benchmark. Furthermore, comprehensive ablation studies validate the critical contributions and the strong synergistic effects of each designed component.

{Despite these promising advancements, certain limitations warrant critical reflection. As noted in Section 4.6, the model exhibits average performance on extremely fine-grained attributes, such as ``glasses" or ``backpack". The potential reason for this limitation is twofold. First, event cameras primarily respond to dynamic pixel intensity changes; therefore, small and relatively static objects (e.g., glasses) may not generate sufficient event signals compared to major body movements (e.g., swinging arms or legs). Second, current cross-attention and memory retrieval mechanisms are predominantly optimized for global feature representations, which might inadvertently suppress localized fine-grained visual cues.} Additionally, the multi-stage nature of the architecture introduces a degree of sensitivity to hyperparameter settings.

{To address these weaknesses, future research will focus on several key directions. First, to overcome the bottleneck in fine-grained attribute recognition, future iterations could incorporate local region-aware attention mechanisms or part-based memory banks to explicitly capture small-scale details. Second, efforts will be directed towards simplifying the complex multi-stage pipeline into a unified, end-to-end trainable architecture, thereby reducing hyperparameter sensitivity. Finally, extending the proposed frequency-domain prompting strategy to other resource-constrained multimodal perception tasks remains a highly promising avenue for future exploration.}

\section*{Acknowledgments}
The authors are grateful to the support of the Guangdong Key Disciplines Project (2024ZDJS137).

\bibliographystyle{apalike}
\bibliography{reference}

\begin{thebibliography}{}

\bibitem[Abdulnabi et~al., 2015]{Abdulnabi}
Abdulnabi, A.~H., Wang, G., Lu, J., and Jia, K. (2015).
\newblock Multi-task cnn model for attribute prediction.
\newblock {\em IEEE Transactions on Multimedia}, 17(11):1949–1959.

\bibitem[Bommasani and et~al.,
  2022]{bommasani2022opportunitiesrisksfoundationmodels}
Bommasani, R. and et~al., D. A.~H. (2022).
\newblock On the opportunities and risks of foundation models.

\bibitem[Cao et~al., 2023]{cao2023novel}
Cao, Y., Fang, Y., Zhang, Y., Hou, X., Zhang, K., and Huang, W. (2023).
\newblock A novel self-boosting dual-branch model for pedestrian attribute
  recognition.
\newblock {\em Signal Processing: Image Communication}, 115:116961.

\bibitem[Chen et~al., 2026]{chen2026subjective}
Chen, W., Ye, C., Song, P., Zhang, Y., and Mao, Z. (2026).
\newblock Subjective-objective emotion correlated generation network for
  subjective video captioning.
\newblock {\em IEEE Transactions on Image Processing}.

\bibitem[Cheng et~al., 2022]{VTB}
Cheng, X., Jia, M., Wang, Q., and Zhang, J. (2022).
\newblock A simple visual-textual baseline for pedestrian attribute
  recognition.
\newblock {\em IEEE Transactions on Circuits and Systems for Video Technology},
  32(10):6994--7004.

\bibitem[Croitoru et~al., 2023]{croitoru2023diffusion}
Croitoru, F.-A., Hondru, V., Ionescu, R.~T., and Shah, M. (2023).
\newblock Diffusion models in vision: A survey.
\newblock {\em IEEE transactions on pattern analysis and machine intelligence},
  45(9):10850--10869.

\bibitem[Dalal and Triggs, 2005]{HOG}
Dalal, N. and Triggs, B. (2005).
\newblock Histograms of oriented gradients for human detection.
\newblock In {\em 2005 IEEE Computer Society Conference on Computer Vision and
  Pattern Recognition (CVPR'05)}, volume~1, pages 886--893 vol. 1.

\bibitem[Fan et~al., 2024]{fan2024parformer}
Fan, X., Zhang, Y., Lu, Y., and Wang, H. (2024).
\newblock Parformer: Transformer-based multi-task network for pedestrian
  attribute recognition.
\newblock {\em IEEE Trans. Cir. and Sys. for Video Technol.}, 34(1):411–423.

\bibitem[Gallego et~al., 2020]{gallego2020eventSurvey}
Gallego, G., Delbr{\"u}ck, T., Orchard, G., Bartolozzi, C., Taba, B., Censi,
  A., Leutenegger, S., Davison, A.~J., Conradt, J., Daniilidis, K., et~al.
  (2020).
\newblock Event-based vision: A survey.
\newblock {\em IEEE transactions on pattern analysis and machine intelligence},
  44(1):154--180.

\bibitem[Hu et~al., 2025]{Hu}
Hu, Y., Chen, X., Wang, S., Liu, L., Shi, H., Fan, L., Tian, J., and Liang, J.
  (2025).
\newblock Deformable cross-attention transformer for weakly aligned rgb–t
  pedestrian detection.
\newblock {\em IEEE Transactions on Multimedia}, 27:4400--4411.

\bibitem[Huang et~al., 2024]{huang2024attributeRetrieval}
Huang, Y., Zhang, Z., Wu, Q., Zhong, Y., and Wang, L. (2024).
\newblock Attribute-guided pedestrian retrieval: Bridging person re-id with
  internal attribute variability.
\newblock In {\em Proceedings of the IEEE/CVF conference on computer vision and
  pattern recognition}, pages 17689--17699.

\bibitem[Jia et~al., 2021]{jia2021spatial}
Jia, J., Chen, X., and Huang, K. (2021).
\newblock Spatial and semantic consistency regularizations for pedestrian
  attribute recognition.

\bibitem[Jia et~al., 2020]{rethinkingofpar}
Jia, J., Huang, H., Yang, W., Chen, X., and Huang, K. (2020).
\newblock Rethinking of pedestrian attribute recognition: Realistic datasets
  with efficient method.
\newblock {\em CoRR}, abs/2005.11909.

\bibitem[Jin et~al., 2026]{jinsequencePAR}
Jin, J., Wang, X., Lin, Y., Li, C., Huang, L., Zheng, A., and Tang, J. (2026).
\newblock Sequencepar: Understanding pedestrian attributes via a sequence
  generation paradigm.
\newblock {\em Pattern Recognition}, 172:112356.

\bibitem[Krotov and Hopfield, 2016]{krotov2016dense}
Krotov, D. and Hopfield, J.~J. (2016).
\newblock Dense associative memory for pattern recognition.

\bibitem[Lai et~al., 2025]{MambaVT}
Lai, S., Liu, C., Zhu, J., Kang, B., Liu, Y., Wang, D., and Lu, H. (2025).
\newblock Mambavt: Spatio-temporal contextual modeling for robust rgb-t
  tracking.
\newblock {\em IEEE Transactions on Circuits and Systems for Video Technology},
  35(9):9312--9323.

\bibitem[Li et~al., 2015]{deepmar}
Li, D., Chen, X., and Huang, K. (2015).
\newblock Multi-attribute learning for pedestrian attribute recognition in
  surveillance scenarios.
\newblock In {\em 2015 3rd IAPR Asian Conference on Pattern Recognition
  (ACPR)}, pages 111--115.

\bibitem[Li et~al., 2018]{8486604}
Li, D., Chen, X., Zhang, Z., and Huang, K. (2018).
\newblock Pose guided deep model for pedestrian attribute recognition in
  surveillance scenarios.
\newblock In {\em 2018 IEEE International Conference on Multimedia and Expo
  (ICME)}, pages 1--6.

\bibitem[Li et~al., 2025]{li2025SAFE}
Li, D., Jin, J., Zhang, Y., Zhong, Y., Wu, Y., Chen, L., Wang, X., and Luo, B.
  (2025).
\newblock Semantic-aware frame-event fusion based pattern recognition via large
  vision--language models.
\newblock {\em Pattern Recognition}, 158:111080.

\bibitem[Li et~al., 2022]{label2label}
Li, W., Cao, Z., Feng, J., Zhou, J., and Lu, J. (2022).
\newblock Label2label: A language modeling framework for multi-attribute
  learning.

\bibitem[Li et~al., 2024]{li2024attmot}
Li, Y., Xiao, Z., Yang, L., Meng, D., Zhou, X., Fan, H., and Zhang, L. (2024).
\newblock Attmot: improving multiple-object tracking by introducing auxiliary
  pedestrian attributes.
\newblock {\em IEEE transactions on neural networks and learning systems},
  36(3):5454--5468.

\bibitem[Mao et~al., 2025]{mao2025realcustom++}
Mao, Z., Huang, M., Ding, F., Liu, M., He, Q., and Zhang, Y. (2025).
\newblock Realcustom++: Representing images as real textual word for real-time
  customization.
\newblock {\em IEEE Transactions on Pattern Analysis and Machine Intelligence}.

\bibitem[Mao et~al., 2026]{11358752}
Mao, Z., Huang, M., Lin, Y., Wang, Q., Zhang, L., and Zhang, Y. (2026).
\newblock Toward accurate image generation via dynamic generative image
  transformer.
\newblock {\em IEEE Transactions on Pattern Analysis and Machine Intelligence},
  pages 1--18.

\bibitem[Ramsauer et~al., 2021]{ramsauer2021hopfieldnetworksneed}
Ramsauer, H., Schäfl, B., Lehner, J., Seidl, P., Widrich, M., Adler, T.,
  Gruber, L., Holzleitner, M., Pavlović, M., Sandve, G.~K., Greiff, V., Kreil,
  D., Kopp, M., Klambauer, G., Brandstetter, J., and Hochreiter, S. (2021).
\newblock Hopfield networks is all you need.

\bibitem[Ristani et~al., 2016]{duke}
Ristani, E., Solera, F., Zou, R., Cucchiara, R., and Tomasi, C. (2016).
\newblock Performance measures and a data set for multi-target, multi-camera
  tracking.
\newblock In {\em European conference on computer vision}, pages 17--35.
  Springer.

\bibitem[Sarafianos et~al., 2018]{sarafianos2018deep}
Sarafianos, N., Xu, X., and Kakadiaris, I.~A. (2018).
\newblock Deep imbalanced attribute classification using visual attention
  aggregation.
\newblock In {\em Proceedings of the European conference on computer vision
  (ECCV)}, pages 680--697.

\bibitem[Schuhmann et~al., 2022]{schuhmann2022laion5bopenlargescaledataset}
Schuhmann, C., Beaumont, R., Vencu, R., Gordon, C., Wightman, R., Cherti, M.,
  Coombes, T., Katta, A., Mullis, C., Wortsman, M., Schramowski, P., Kundurthy,
  S., Crowson, K., Schmidt, L., Kaczmarczyk, R., and Jitsev, J. (2022).
\newblock Laion-5b: An open large-scale dataset for training next generation
  image-text models.

\bibitem[Song et~al., 2026a]{SONG2026113420}
Song, C., Liu, X., Hui, C., Zhu, H., Mi, Y., Geng, K., Wu, J., Zhou, Z., and
  Jiang, F. (2026a).
\newblock Affective-aware fine-grained image quality assessment via multi-modal
  large language models.
\newblock {\em Pattern Recognition}, 178:113420.

\bibitem[Song et~al., 2026b]{SONG2026113278}
Song, X., Zhang, X., Zhao, Q., Wei, B., Guo, X., and Hei, X. (2026b).
\newblock Depth correction and edge guidance network for rgb-d salient object
  detection.
\newblock {\em Pattern Recognition}, 176:113278.

\bibitem[Tang et~al., 2019]{tang2019improving}
Tang, C., Sheng, L., Zhang, Z., and Hu, X. (2019).
\newblock Improving pedestrian attribute recognition with weakly-supervised
  multi-scale attribute-specific localization.
\newblock In {\em Proceedings of the IEEE International Conference on Computer
  Vision}, pages 4997--5006.

\bibitem[Tu et~al., 2025]{MAPNet}
Tu, Z., Qian, X., and Zhou, W. (2025).
\newblock Efficient rgb-d co-salient object detection via modality-aware
  prompting.
\newblock {\em IEEE Transactions on Automation Science and Engineering},
  22:12911--12921.

\bibitem[Wang et~al., 2016]{wang2016cnnrnnunifiedframeworkmultilabel}
Wang, J., Yang, Y., Mao, J., Huang, Z., Huang, C., and Xu, W. (2016).
\newblock Cnn-rnn: A unified framework for multi-label image classification.

\bibitem[Wang et~al., 2025a]{wang2025HARDVS}
Wang, S., Wang, X., Jiang, B., Zhu, L., Li, G., Wang, Y., Tian, Y., and Tang,
  J. (2025a).
\newblock Human activity recognition using rgb-event based sensors: A
  multi-modal heat conduction model and a benchmark dataset.
\newblock {\em arXiv preprint arXiv:2504.05830}.

\bibitem[Wang et~al., 2023]{wang2023PTMSurvey}
Wang, X., Chen, G., Qian, G., Gao, P., Wei, X.-Y., Wang, Y., Tian, Y., and Gao,
  W. (2023).
\newblock Large-scale multi-modal pre-trained models: A comprehensive survey.
\newblock {\em Machine Intelligence Research}, 20(4):447--482.

\bibitem[Wang et~al., 2024a]{PROMPTPAR}
Wang, X., Jin, J., Li, C., Tang, J., Zhang, C., and Wang, W. (2024a).
\newblock Pedestrian attribute recognition via clip based prompt
  vision-language fusion.

\bibitem[Wang et~al., 2024b]{MvHeat-DET}
Wang, X., Jin, Y., Wu, W., Zhang, W., Zhu, L., Jiang, B., and Tian, Y. (2024b).
\newblock Object detection using event camera: A moe heat conduction based
  detector and a new benchmark dataset.

\bibitem[Wang et~al., 2024c]{mambapar}
Wang, X., Kong, W., Jin, J., Wang, S., Gao, R., Ma, Q., Li, C., and Tang, J.
  (2024c).
\newblock An empirical study of mamba-based pedestrian attribute recognition.

\bibitem[Wang et~al., 2025b]{wang2025EvSLT}
Wang, X., Li, Y., Wang, F., Jiang, B., Wang, Y., Tian, Y., Tang, J., and Luo,
  B. (2025b).
\newblock Sign language translation using frame and event stream: Benchmark
  dataset and algorithms.
\newblock {\em arXiv preprint arXiv:2503.06484}.

\bibitem[Wang et~al., 2025c]{AMMRG}
Wang, X., Wang, F., Wang, H., Jiang, B., Li, C., Wang, Y., Tian, Y., and Tang,
  J. (2025c).
\newblock Activating associative disease-aware vision token memory for
  llm-based x-ray report generation.

\bibitem[Wang et~al., 2025d]{EventPAR}
Wang, X., Wang, H., Wang, S., Chen, Q., Jin, J., Song, H., Jiang, B., and Li,
  C. (2025d).
\newblock Rgb-event based pedestrian attribute recognition: A benchmark dataset
  and an asymmetric rwkv fusion framework.

\bibitem[Wang et~al., 2024d]{MaHDFT}
Wang, X., Wang, S., Ding, Y., Li, Y., Wu, W., Rong, Y., Kong, W., Huang, J.,
  Li, S., Yang, H., Wang, Z., Jiang, B., Li, C., Wang, Y., Tian, Y., and Tang,
  J. (2024d).
\newblock State space model for new-generation network alternative to
  transformers: A survey.

\bibitem[Wang et~al., 2022]{wangsurvey}
Wang, X., Zheng, S., Yang, R., Zheng, A., Chen, Z., Tang, J., and Luo, B.
  (2022).
\newblock Pedestrian attribute recognition: A survey.
\newblock {\em Pattern Recognition}, 121:108220.

\bibitem[Xiong et~al., 2026]{XIONG2026113333}
Xiong, J., Xie, X., Feng, Q., and Lai, J.-H. (2026).
\newblock Clip -powered modality centering with spiral training for
  visible-infrared person re-identification.
\newblock {\em Pattern Recognition}, 177:113333.

\bibitem[Xue et~al., 2026]{FMTrack}
Xue, Y., Jin, G., Zhong, B., Shen, T., Tan, L., Xue, C., and Zheng, Y. (2026).
\newblock Fmtrack: Frequency-aware interaction and multi-expert fusion for
  rgb-t tracking.
\newblock {\em IEEE Transactions on Circuits and Systems for Video Technology},
  36(2):1655--1667.

\bibitem[Ye et~al., 2025]{ye2025improving}
Ye, C., Chen, W., Hu, B., Zhang, L., Zhang, Y., and Mao, Z. (2025).
\newblock Improving video summarization by exploring the coherence between
  corresponding captions.
\newblock {\em IEEE Transactions on Image Processing}.

\bibitem[Yuan et~al., 2023]{HAP}
Yuan, J., Zhang, X., Zhou, H., Wang, J., Qiu, Z., Shao, Z., Zhang, S., Long,
  S., Kuang, K., Yao, K., Han, J., Ding, E., Lin, L., Wu, F., and Wang, J.
  (2023).
\newblock Hap: Structure-aware masked image modeling for human-centric
  perception.

\bibitem[Zhao et~al., 2019]{zhao2019recurrent}
Zhao, X., Sang, L., Ding, G., Han, J., Di, N., and Yan, C. (2019).
\newblock Recurrent attention model for pedestrian attribute recognition.
\newblock In {\em Proceedings of the AAAI Conference on Artificial
  Intelligence}, volume~33, pages 9275--9282.

\bibitem[Zheng et~al., 2023]{zheng2023diverse}
Zheng, A., Wang, H., Wang, J., Huang, H., He, R., and Hussain, A. (2023).
\newblock Diverse features discovery transformer for pedestrian attribute
  recognition.
\newblock {\em Engineering Applications of Artificial Intelligence},
  119:105708.

\bibitem[Zhou et~al., 2024a]{zhou2024pedestrian}
Zhou, Y., Hu, H.-M., Xiang, Y., Zhang, X., and Wu, H. (2024a).
\newblock Pedestrian attribute recognition as label-balanced multi-label
  learning.
\newblock {\em arXiv preprint arXiv:2405.04858}.

\bibitem[Zhou et~al., 2024b]{sspnet}
Zhou, Y., Hu, H.-M., Xiang, Y., Zhang, X., and Wu, H. (2024b).
\newblock Pedestrian attribute recognition as label-balanced multi-label
  learning.

\bibitem[Zhou et~al., 2023]{zhou2023}
Zhou, Y., Hu, H.-M., Yu, J., Xu, Z., Lu, W., and Cao, Y. (2023).
\newblock A solution to co-occurrence bias: Attributes disentanglement via
  mutual information minimization for pedestrian attribute recognition.

\bibitem[Zhu et~al., 2023]{zhu2023visual}
Zhu, J., Lai, S., Chen, X., Wang, D., and Lu, H. (2023).
\newblock Visual prompt multi-modal tracking.
\newblock In {\em Proceedings of the IEEE/CVF conference on computer vision and
  pattern recognition}, pages 9516--9526.

\bibitem[Zhu et~al., 2025]{zhu2025crsot}
Zhu, Y., Wang, X., Li, C., Jiang, B., Zhu, L., Huang, Z., Tian, Y., and Tang,
  J. (2025).
\newblock Crsot: Cross-resolution object tracking using unaligned frame and
  event cameras.
\newblock {\em IEEE Transactions on Multimedia}.

\bibitem[Zuo et~al., 2024]{zuo2024plip}
Zuo, J., Hong, J., Zhang, F., Yu, C., Zhou, H., Gao, C., Sang, N., and Wang, J.
  (2024).
\newblock Plip: Language-image pre-training for person representation learning.

\end{thebibliography}


\end{document}